\documentclass[letterpaper, 10 pt, conference]{ieeeconf}  

\IEEEoverridecommandlockouts                              

\usepackage{graphics} 
\usepackage{times} 
\usepackage{amsmath} 
\usepackage{amssymb}  
\usepackage[labelformat=simple]{subcaption}

\usepackage{tikz}
\usetikzlibrary{positioning}
\newcommand{\nsp}{\hspace{-0.5mm}}
\newcommand{\kl}{D_{\hspace{-0.2mm}K\hspace{-0.2mm}L\hspace{-0.2mm}}}
\newcommand{\minus}{\scalebox{0.75}[1.0]{$-$}}
\title{\LARGE \bf
Learning and Predicting Multimodal Vehicle Action Distributions in a Unified Probabilistic
Model Without Labels}

\author{Charles Richter, Patrick R. Barrag\'{a}n*, and Sertac Karaman*
\thanks{*Co-advised work. The authors are affiliated with Optimus Ride.
\texttt{\scriptsize\{charlie,patrick,sertac\}@optimusride.com}. Presented at the Fresh Perspectives
on the Future of Autonomous Driving workshop, ICRA 2022.}%
}

\begin{document}

\maketitle
\thispagestyle{empty}
\pagestyle{empty}

\begin{abstract}
  We present a unified probabilistic model that learns a representative set of discrete vehicle
  actions and predicts the probability of each action given a particular scenario. Our model also
  enables us to estimate the distribution over continuous trajectories conditioned on a scenario,
  representing what each discrete action would look like if executed in that scenario. While our
  primary objective is to learn representative action sets, these capabilities combine to produce
  accurate multimodal trajectory predictions as a byproduct. Although our learned action
  representations closely resemble semantically meaningful categories (e.g., ``go straight'', ``turn
  left'', etc.), our method is entirely self-supervised and does not utilize any manually generated
  labels or categories. Our method builds upon recent advances in variational inference and deep
  unsupervised clustering, resulting in full distribution estimates based on deterministic model
  evaluations.
\end{abstract}

\IEEEpeerreviewmaketitle

\section{Introduction}

A central challenge in robotics and artificial intelligence is to develop discrete representations
that can translate the high-dimensional continuous spaces of real-world sensor data and robot
configuration into forms that are compatible with algorithms for abstract reasoning, such as search
and logical or probabilistic inference~\cite{konidaris2019necessity}. Although representation
learning has been studied extensively in the machine learning
literature~\cite{bengio2013representation}, and learned action representations are often used in
robotics, it remains an open challenge to distill unlabeled natural data into a representative set
of discrete actions. In particular, learned discrete action representations have not been widely
adopted in the recent autonomous vehicle literature, possibly because many essential components of
an autonomous vehicle system can be engineered or learned to a considerable degree without them. For
example, predicted trajectories for other vehicles on the road can be fed directly into a planning
system to avoid collision, without those predicted trajectories representing distinct maneuvers.

Nevertheless, there are important cases in an autonomous vehicle system in which it would be useful
to describe behavior in terms of a representative discrete action set. For example, communicating
intent or receiving instruction from a user would require a relatively small set of meaningfully
distinct action choices, perhaps with semantic labels attached. Similarly, right-of-way conventions
and rules of the road are understood in terms of discrete actions, necessitating a way to classify
continuous-valued trajectories as members of an action set in order to evaluate their legality.

At the same time, we observe that there is enormous variability in the behaviors of drivers in
real-world scenarios. There is also great diversity in the design of roads, intersections, parking
lots, sidewalks and other environments where we must understand and predict movements of other
agents. There is a substantial challenge in developing a representative discrete action set under
these highly variable conditions. For example, if a vehicle is following a road that curves gently
to the left, does that constitute a ``go straight'' or ``turn left'' maneuver? We adopt the position
that manually defining maneuvers and classifiers beyond the few simplest categories will quickly
become untenable and therefore the categories themselves should arise automatically from the data
without manually generated labels.

Our goal in this paper is primarily to learn a representative discrete set of actions that can be
used to describe and predict intent, and secondarily to forecast continuous-valued trajectories
using that set. We aim to provide a means of answering queries such as ``What is the probability of
taking action $i$ in this scenario?'', as well as: ``What would the distribution of action $i$ look
like if executed in this scenario?'' An additional goal is to estimate full distributional
information, since uncertainty information is essential to safe and effective autonomous vehicle
operation. To achieve these objectives, we develop a unified probabilistic model based on methods
that combine variational inference and unsupervised clustering. Our model assumes that trajectories
are explained by both discrete and continuous underlying latent variables, and that those underlying
factors are determined by (or can be inferred from) an input scenario. We begin by motivating our
work with respect to relevant literature, and then we develop our model and illustrate its
effectiveness in both learning and predicting vehicle motions.

\section{Related Work}

Recently, the autonomous vehicle literature has proliferated with many successful motion prediction
methods enabled by publicly available motion datasets and advances in machine learning. Most methods
formulate their input as a rasterized birds-eye-view (BEV) image containing semantic road features
(lane lines, traffic signals, crosswalks, etc.) along with additional channels containing past
states of agents in the scene. This image can be fed as input to a CNN to extract features and
predict the future trajectory of the target of interest. Examples
include~\cite{Hong2019RulesOT,cui2019multimodal,chai2019multipath,phan2020covernet} while others
augment the rasterized input with raw sensor data~\cite{casas2018intentnet}. Although most methods
predict individual agent motions using a CNN, some explicitly reason about multi-agent interaction
~\cite{lee2017desire,tang2019multiple,casas2020implicit} and utilize other learning methods such as
recurrent and graph neural networks~\cite{mozaffari2020deep}.

Some methods frame multi-modal prediction as classification over a set of possible trajectories.
These trajectory sets may consist of samples (or cluster means) from the training
dataset~\cite{chai2019multipath}, or may be generated based on known
dynamics~\cite{phan2020covernet}. While effective for continuous motion prediction, these action
sets are often too large and dense to form a parsimonious representation for logical or semantic
reasoning, and they may not be optimized to model the true data distribution. In contrast, our
method optimizes a much smaller set of action \emph{distributions} by maximizing data likelihood
under a unified probabilistic model.

Other methods predict discrete intent or behavior, sometimes along with a motion prediction. Many of
these methods assume a set of manually-defined categories for simple maneuvers like ``go straight''
or ``turn left'' \cite{7995948,casas2018intentnet,deo2018convolutional}.  Those that do not manually
define behavior categories often instead limit their scope to very specific road structures like
3-way and 4-way intersections~\cite{zhang2013understanding} or limit their evaluation to very
specific scenarios like highway driving which lack the full diversity of vehicle
maneuvers~\cite{hu2018probabilistic}. In this work, we do not use any manually designed maneuver
categories or labels, and we make no assumptions about roadway structure or vehicle behaviors so
that the action representation we learn will be as general and representative as possible.

To capture variability in vehicle or pedestrian behavior, some methods have utilized latent variable
models such as conditional variational autoencoders ((C)VAEs). A discrete latent variable is often
used to capture multimodality~\cite{tang2019multiple,ivanovic2020multimodal,Hong2019RulesOT}, while
a continuous latent variable can be used to capture general variability in a distribution of
actions~\cite{lee2017desire,casas2020implicit}. Our work differs from these and is unique in that we
utilize \emph{both} discrete and continuous latent variables. The discrete latent variable in our
model captures multimodality in the space of behaviors, while the continuous latent variable allows
diversity within each behavior, so that actions can be adapted and shaped to the specific context as
appropriate. \cite{rakos2020compression} use a continuous VAE encoding to classify different
manually-defined maneuvers, but do not use a discrete latent variable to identify and cluster those
maneuvers as we do. Other latent variable methods have been developed to draw diverse samples that
represent distinct motion behavior modes in CVAE~\cite{yuan2019diverse} and
GAN~\cite{huang2020diversitygan} models.

Self-supervised learning of a representative action set can be viewed as a clustering problem, and
our approach is partially inspired by recent methods that combine clustering with deep learning and
variational inference. Since the introduction of VAEs~\cite{kingma2013auto}, many subsequent works
have replaced the unimodal Gaussian latent space prior with more complex
distributions~\cite{jang2016categorical,joo2020dirichlet,tomczak2018vae,nalisnick2016stick,rezende2015variational}.
The Gaussian Mixture Model (GMM) latent distribution in particular has been used to facilitate deep
unsupervised clustering of data like handwritten digits (e.g., MNIST), where elements have a
discrete identity as well as continuous style
variation~\cite{dilokthanakul2016deep,ijcai2017-273,rao2019continual}. With a GMM latent
distribution, a (C)VAE optimizes a set of Gaussian clusters in the jointly learned continuous latent
space. We use exactly this approach to encode and cluster vehicle motion data into discrete actions.

\section{Learning and Predicting Multi-Modal Action Distributions}

Our primary objective is to jointly learn a set of representative actions and a model that predicts
the discrete probability of each action in a given scenario. We consider an action to be not a
single fixed trajectory, but a continuous distribution of similar trajectories that might serve the
same functional purpose across a range of different scenarios and map geometries. Therefore, we have
both discrete and continuous elements of variation that we wish to model, however we assume no prior
knowledge of manually-defined behavior categories or action shapes and we aim for both the discrete
action categories and the associated continuous-valued distributions over trajectories to arise
naturally from the data. First, we will develop a model that learns a representative set of actions
in a self-supervised manner, and then we will extend that model to make accurate motion predictions
conditioned on a given scenario.

\subsection{Learning a Set of Actions}
\label{sec:actionset}

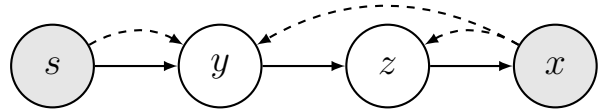
\begin{figure}
\centering
\begin{tikzpicture}
\tikzstyle{main}=[circle,minimum size=11mm,thick,draw=black!100,node distance=11mm]
\tikzstyle{connect}=[-latex, thick]
  \node[main, fill = black!10] (s) [label=center:{\Large$s$}] { };
  \node[main] (y) [right=of s,label=center:{\Large$y$}] { };
  \node[main] (z) [right=of y,label=center:{\Large$z$}] {};
  \node[main, fill = black!10] (x) [right=of z,label=center:{\Large$x$}] { };
  \path (s) edge [connect] (y)
        (y) edge [connect] (z)
				(z) edge [connect] (x)
				(x) edge [connect, bend right, dashed] (z)
				(x) edge [connect, bend right, dashed] (y)
				(s) edge [connect, bend left, dashed] (y);
\end{tikzpicture}
  \caption{Graphical model in which observed variables $s$ and $x$ are shaded while latent variables
  $y$ and $z$ are not shaded. Solid arrows indicate the generative model while dashed arrows
  indicate the variational model.}
  \label{model}
\end{figure}

We assume a dataset of trajectory-scenario pairs $(X,S) = ((x_1,s_1),\ldots,(x_N,s_N))$, where $x_i$
is a vector of future vehicle position coordinates for the target vehicle of interest in data sample
$i$, and $s_i$ is a scenario (or context) represented as a multi-channel rasterization of the map
and past states of the target vehicle and other agents in the scenario. To capture both discrete and
continuous elements of variation in driving behaviors, we propose a latent variable model using both
discrete and continuous latent variables. We model the system according to the graphical model
illustrated in Figure~\ref{model}. In this model, $x$ and $s$ represent the trajectory and scenario,
respectively, $y\in{1,\ldots,K}$ is a discrete latent variable and $z\in\mathbb{R}^D$ is a
continuous latent variable. The latent variables $y$ and $z$ together constitute a Gaussian mixture
model (GMM) where the value of $y$ selects a particular component of the GMM and each component is
represented with a mean and covariance. Our goal is to maximize $\text{ln }p(X|S)$ over the
parameters of this model, thereby learning a set of GMM components that decode via $p(x|z)$ to
continuous action distributions, as well as a predictor $p(y|s)$ of discrete action probabilities
given a scenario.

Since direct optimization of this model is difficult, we adopt the method of variational inference
and specify a variational model $q(y,z|x,s)$ to approximate the distribution over latent variables.
We assume that the generative and variational models factorize as follows:
\begin{align}
  p(x,y,z|s) &= p(y|s)p(z|y)p(x|z)\\
  q(y,z|x,s) &= q(y|x,s)q(z|x).
\end{align}
In Figure~\ref{model}, the generative model is illustrated with solid arrows while the variational
model is illustrated with dashed arrows. This factorization gives rise to a relationship between $z$
and $x$ that closely resembles a conventional variational autoencoder, where $q(z|x)$ acts as an
encoder into a continuous-valued latent space and $p(x|z)$ acts as a decoder. However, unlike a
conventional VAE, our latent space prior comprises $K$ Gaussians whose parameters are optimized to
capture the modes of the encoded data. Learning in this model simultaneously shapes the latent space
$z$ and clusters data in that space. These clusters, when decoded to the trajectory space $x$ define
our learned action set.

Following a common decomposition in variational inference, the log likelihood for a single element
of our dataset can be written as:
\begin{equation}
  \text{ln }p(x|s) = \mathcal{L}(q) + \kl(q||p),
\end{equation}
where:
\begin{equation}
  \mathcal{L}(q) = \sum_y\int_z q(y,z|x,s)\text{ln}\left\{\frac{p(x,y,z|s)}{q(y,z|x,s)}\right\}dz
  \label{elbo_unfactored}
\end{equation}
and:
\begin{equation}
  \kl(q||p) = \minus\sum_y\int_z q(y,z|x,s)\text{ln}\left\{\frac{p(y,z|x,s)}{q(y,z|x,s)}\right\}dz.
\end{equation}
Since $\kl(q||p)$ is always non-negative, we can use $\mathcal{L}(q)$ as a lower bound on the data
likelihood, known as the evidence lower bound (ELBO). Optimizing this bound $\mathcal{L}(q)$
equivalently maximizes data likelihood and minimizes $\kl(q||p)$. Substituting our factorization of
$p(x,y,z|s)$ and $q(y,z|x,s)$ into equation~\eqref{elbo_unfactored} gives:
\begin{align}
  \hspace{-1mm}\mathcal{L} &= \sum_y\int_z q(y|x,s)q(z|x)
    \text{ln}\left\{\frac{p(y|s)p(z|y)p(x|z)}{q(y|x,s)q(z|x)}\right\}dz\label{elbo_0}\\
  \begin{split}\label{elbo}
  &=\int_z q(z|x)\text{ln }p(x|z)dz\minus\kl\left(q(y|x,s)||p(y|s)\right)\\
	&\qquad\minus\sum_y q(y|x,s)\kl\left(q(z|x)||p(z|y)\right).
  \end{split}
\end{align}
We implement $p(y|s)$, $p(x|z)$ and $q(z|x)$ as neural networks and we implement $p(z|y)$ as a
linear function mapping a one-of-$K$ representation of $y$ to mean and variance of clusters in
latent space $z$.  While $p(x|z)$ and $q(z|x)$ are fully connected networks, $p(y|s)$ includes a
convolutional stage that extracts features from the rasterized scene input. As in many VAE
implementations, the encoder $q(z|x)$ learns both means and variances of $z$, while the decoder
$p(x|z)$ learns only means of $x$ and the output distribution is defined as a Gaussian with identity
covariance (though that variance could also be learned). Note that the variance of $z$, not $x$, is
what induces the distributional spread of trajectories within an action.

We do not represent $q(y|x,s)$ as a neural network, since this distribution can be directly computed
using the other distributions. This particular model and factorization follow a mean field
approximation since the variational approximation for neither $y$ nor $z$ depends on the other,
therefore we follow a general result~\cite{10.5555/1162264} to obtain (see
Appendix~\ref{appendix:derivations}):
\begin{equation}
  q(y|x,s) = \frac{p(y|s)\text{exp}(\minus H(q(z|x),p(z|y)))}{\sum_y p(y|s)\text{exp}(\minus
  H(q(z|x),p(z|y)))},
  \label{qy}
\end{equation}
where $H(q(z|x),p(z|y))$ is the cross entropy between $q(z|x)$ and $p(z|y)$, which can be computed
analytically since both distributions are Gaussian. This expression provides the intuitive result
that probabilistic assignment of clusters is determined by proximity between the encoding given by
$q(z|x)$ and the cluster location given by $p(z|y)$. Computing $q(y|x,s)$ in this way is analogous
to computing the E-step in the expectation maximization algorithm.

For the purposes of optimizing equation~\ref{elbo} with stochastic gradient descent, we can
approximate the integral in the first term with a Monte Carlo estimate using a single sample,
leading to the objective function:
\begin{equation}
  \begin{split}\label{loss}
    \mathcal{L} \approx \text{ln }p(x|\tilde{z})\minus\kl\left(q(y|x,s)||p(y|s)\right)\\
	  \minus\sum_y q(y|x,s)\kl\left(q(z|x)||p(z|y)\right),
	\end{split}
\end{equation}
where $\tilde{z}$ is a sample drawn from $q(z|x)$. At each training step, we compute $q(y|x,s)$
using equation~\eqref{qy} with the current model parameter values and hold that distribution fixed
while we optimize the other distributions.

\begin{figure}
  \centering
  \includegraphics[width=\linewidth]{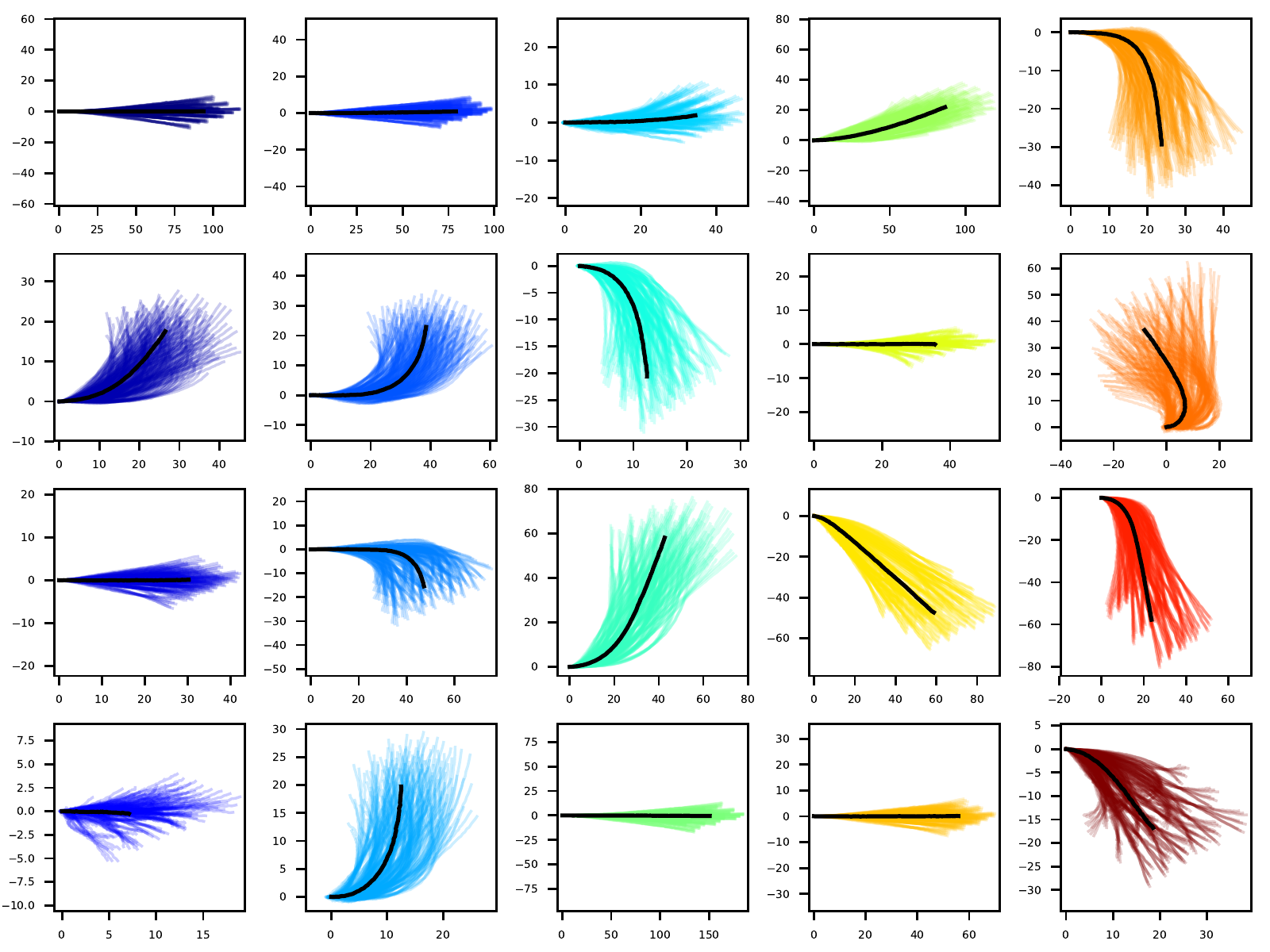}
  \caption{Learned action distributions with latent space means illustrated in black and $\pm
    1\sigma$ points illustrated in color. All axes represent distance in meters.}
  \label{fig:action_distribution}
\end{figure}

\begin{figure}
  \centering
  \includegraphics[width=\linewidth,trim=70 115 130 65,clip]{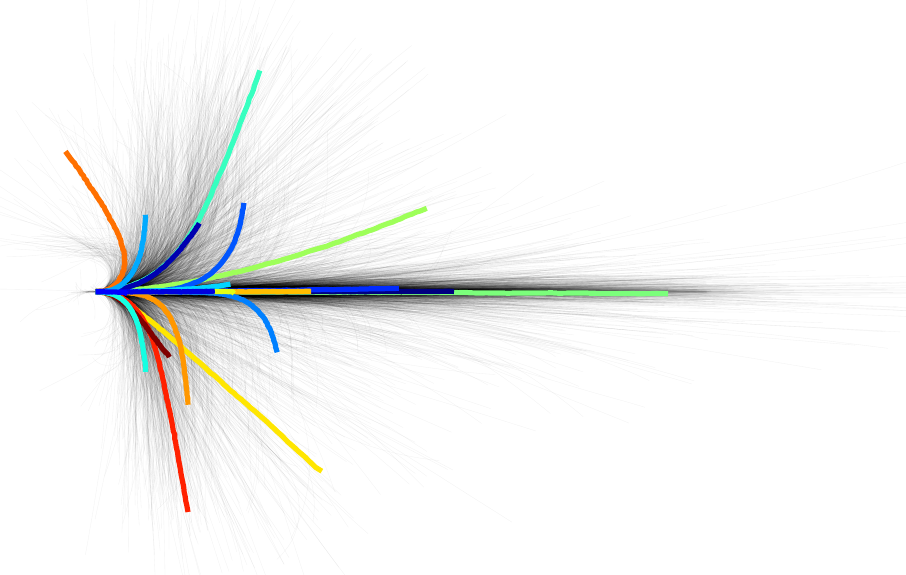}
  \caption{Learned action means decoded into trajectory space, overlaid on trajectory samples from
  the dataset.}
  \label{fig:action_overlay}
\end{figure}

Figure~\ref{fig:action_distribution} illustrates a set of actions learned using this method. These
actions include a diverse range, both turning and straight, that differ in shape and vehicle speed
(and therefore scale). While the mean of each distribution (illustrated in black) shows the overall
characteristic of the action, the distribution around it (illustrated in color) shows the
variability within each action. The mean is illustrated by decoding the mean value of each GMM
cluster component, while the distributions are illustrated by decoding points that are offset from
the mean by $\pm 1\sigma$ in each latent dimension. We note that these samples are generated in a
single deterministic evaluation of the decoder on a tensor of sigma points without the need for
random sampling.  Figure~\ref{fig:action_overlay} illustrates the mean of each action distribution
overlaid on trajectory samples from the Waymo motion dataset~\cite{ettinger2021large}, showing
strong qualitative agreement between the learned actions and the types of trajectories in the
dataset.

\subsection{Accurate Predictions with Learned Action Distributions}
\label{sec:prediction}

In Section~\ref{sec:actionset}, we developed a model to learn a set of representative action
distributions and predict the discrete probability $p(y|s)$ of each action being selected given a
particular scenario. However, since these action distributions are not yet conditioned on that
scenario, they are not directly applicable for making accurate motion predictions.
Figures~\ref{fig:f1} and~\ref{fig:f3} show imprecise predictions made by first evaluating $p(y|s)$
to determine likely values of $y$, and then for each corresponding GMM component, decoding values of
$z$ equal to the component mean $\pm 1\sigma$ to obtain a distributional spread of trajectories.
Although the discrete action choices in these examples may be reasonable, the trajectory predictions
are not tailored to the specific scenarios.

More concretely, in the model we developed in~\ref{sec:actionset} (Figure~\ref{model}), the
distribution over the continuous latent variable $z$ does not depend on scenario $s$ except through
the choice of $y$. In reality, the value of $z$ should depend \emph{strongly} on $s$ because it is
the geometric structure of the road and surrounding agents in the scenario that determine which
actions are both feasible and likely. For instance, small variations in the value of $z$ within a
given action distribution make the difference between a turn into the correct lane and a turn into
oncoming traffic.  Therefore, why not introduce a dependency on $s$ in the prediction of the
continuous latent variable, e.g., $p(z|y,s)$ or the decoder, e.g., $p(x|z,s)$?

Since the scenario $s$ contains sufficient information in many cases to accurately predict the
continuous-valued latent variable $z$ or the output trajectory $x$ directly without the use of a
discrete action choice, introducing $s$ as a dependency in these distributions undermines and even
prevents the model from effectively learning a multimodal action representation. Though we do not
quantify this behavior here, we observe that a model whose latent variable prediction has access to
the scenario (e.g., $p(z|y,s)$) will simply bypass the discrete variable altogether, resulting in a
very small number of non-degenerate clusters and one dominant cluster whose distribution spans the
full continuum of actions. Similarly, in a model whose decoder has access to the scenario (e.g.,
$p(x|z,s)$), the geometric shape and semantic role of each discrete action become so shifted
depending on the scenario as to lose any consistent identifiable meaning. We find that in order to
achieve effective clustering of data into representative action distributions for our purposes, and
to learn non-degenerate predictions of $y$ based on $s$, the learned structure of the latent space
must not shift with $s$ and we must have the discrete variable $y$ as the sole conduit or bottleneck
of scenario information in the generative model.

Nevertheless, we must somehow still refine our prediction of the continuous latent variable $z$ in
order to make accurate trajectory predictions. What we are truly interested in is the posterior
$p(z|y,s)$ that can be decoded to generate the distribution of trajectories for a specific discrete
action choice and scenario. Fortunately the methods of variational inference are well suited to
estimating posteriors over latent variables, and our approach will be to approximate the true
posterior $p(z|y,s)$ with a variational model $q(z|y,s)$, effectively serving as a separate encoder.

We could attempt to modify the model illustrated in Figure~\ref{model} by replacing the existing
encoder $q(z|x)$ with a different encoder $q(z|y,s)$, however as we have noted above, such a model
would simply bypass the use of the discrete variable $y$ and simply predict $z$ directly based on
$s$. Therefore, we must maintain the structure illustrated in Figure~\ref{model}, but we explore two
options that build upon that structure. The first option is to train an alternate encoder for our
original model, and the section option is to extend our original model to create a larger unified
model that includes two different encoders, jointly training all of the desired distributions. We
discuss these two possibilities in turn.

\subsection{Learning a Dual Encoder}
\label{sec:dual}

\begin{figure}
\centering
\begin{tikzpicture}
\tikzstyle{main}=[circle,minimum size=11mm,thick,draw=black!100,node distance=11mm]
\tikzstyle{connect}=[-latex, thick]
  \node[main, fill = black!10] (s) [label=center:{\Large$s$}] { };
  \node[main] (y) [right=of s,label=center:{\Large$y$}] { };
  \node[main] (z) [right=of y,label=center:{\Large$z$}] {};
  \node[main, fill = black!10] (x) [right=of z,label=center:{\Large$x$}] { };
  \path (s) edge [connect, draw=black!20] (y)
        (y) edge [connect, draw=black!20] (z)
				(z) edge [connect, draw=black!20] (x)
				(x) edge [connect, bend right, dashed, draw=black!20] (z)
				(x) edge [connect, bend right, dashed, draw=black!20] (y)
				(s) edge [connect, bend left, dashed, draw=black!20] (y)
				(s) edge [connect, bend right, dashed, draw=black!100] (z)
				(y) edge [connect, bend right, dashed, draw=black!100] (z);
\end{tikzpicture}
  \caption{Graphical model in which the alternate encoder to be learned, $q(z|y,s)$, is illustrated
  in black while all other distributions are grey.}
  \label{model_alternate}
\end{figure}
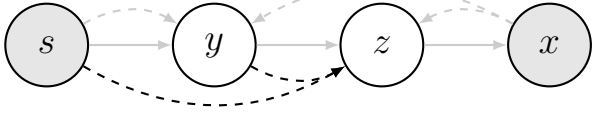

First, we consider learning and fixing in place the distributions in our original model, then
training a separate encoder $q(z|y,s)$ using a variant of our original objective~\eqref{elbo}:
\begin{align}
  \hspace{-1mm}\mathcal{L}\nsp&=\nsp\sum_y\nsp\int_z\nsp q(y|x,\nsp s)q(z|y,\nsp s)
    \text{ln}\nsp\left\{\nsp\frac{p(y|s)p(z|y)p(x|z)}{q(y|x,s)q(z|y,s)}\nsp\right\}\nsp dz\\
  \begin{split}
    &=\nsp\sum_y\nsp q(y|x,\nsp s)\nsp\Biggl[\nsp\int_z\nsp q(z|y,\nsp s) \text{ln }\nsp p(x|z)dz\\[-2mm]
    &\hspace{5mm}\minus\nsp \kl\nsp\left(q(z|y,\nsp s)||p(z|y)\right)\Biggr]\nsp\minus\nsp
    \kl\nsp\left(q(y|x,\nsp s)||p(y|s)\right)
  \end{split}\label{eq:elbo_alternate_expanded}\\
  &\approx\nsp\sum_y\nsp q(y|x,\nsp s)\Bigl[\text{ln }p(x|\tilde{z})\minus\kl\nsp\left(q(z|y,\nsp
  s)||p(z|y)\right)\Bigr]\label{loss_alternate},
\end{align}
where in the final approximation, $\tilde{z}$ is sampled from $q(z|y,s)$ (not $q(z|x)$) separately
for each value of $y$, and where we have dropped the final term from
equation~\eqref{eq:elbo_alternate_expanded} since we assume that $q(y|x,s)$ and $p(y|s)$ are both
known and fixed. Indeed, the objective is to train $q(z|y,s)$ only, and we assume that all other
terms in this function have already been learned and fixed. We illustrate this alternative encoder
in Figure~\ref{model_alternate}.

We consider this to be a VAE-like objective function because it contains the two terms typically
found in a conventional VAE objective function: Reconstruction of the input based on a sample drawn
from the variational posterior $q(z|y,s)$, and the KL divergence between the variational posterior
and the prior. We rely on the discrete distribution $q(y|x,s)$ to correctly attribute each data
sample to the appropriate discrete action, which enables $q(z|y,s)$ to learn to predict different
maneuvers (e.g., ``go straight'' vs. ``turn left'') from the same scenario $s$. We implement
$q(z|y,s)$ as a collection of $K$ neural networks, $q_1(z|s),\ldots,q_K(z|s)$.

Although this method is functional, one particular drawback of learning $q(z|y,s)$ apart from the
rest of the model is that the CNN stage that extracts scene features to predict $p(y|s)$ may not
learn features that are optimally tuned to predicting $q(z|y,s)$. Therefore, it may be necessary to
train a separate convolutional stage specifically for $q(z|y,s)$ or interleave both training phases.
In the next section, we introduce a single unified model that overcomes this problem by jointly
learning all distributions simultaneously.

\subsection{Unified Model}
\label{sec:unified}

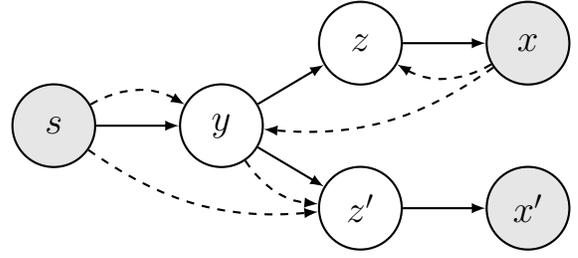
\begin{figure}
\centering
\begin{tikzpicture}
\tikzstyle{main}=[circle,minimum size=11mm,thick,draw=black!100,node distance=11mm]
\tikzstyle{connect}=[-latex, thick]
  \node[main, fill = black!10] (s) [label=center:{\Large$s$}] { };
  \node[main] (y) [right=of s,label=center:{\Large$y$}] { };
  \node[main] (z1) [above right=3mm and 10.5mm of y,label=center:{\Large$z$}] {};
  \node[main] (z2) [below right=3mm and 10.5mm of y,label=center:{\Large$z'$}] {};
  \node[main, fill = black!10] (x1) [right=of z1,label=center:{\Large$x$}] { };
  \node[main, fill = black!10] (x2) [right=of z2,label=center:{\Large$x'$}] { };
  \path (s) edge [connect] (y)
        (y) edge [connect] (z1)
        (y) edge [connect] (z2)
				(z1) edge [connect] (x1)
				(z2) edge [connect] (x2)
				(x1) edge [connect, bend left, dashed] (z1)
				(x1) edge [connect, bend left=20, dashed] (y)
				(s) edge [connect, bend left, dashed] (y)
				(y) edge [connect, bend right=25, dashed] (z2)
				(s) edge [connect, bend right=20, dashed] (z2);
\end{tikzpicture}
  \caption{Unified graphical model.}
  \label{model_unified}
\end{figure}

As we discussed in Sections~\ref{sec:prediction} and \ref{sec:dual}, we wish to learn a set of
representative action distributions and a discrete action predictor, which we can accomplish with a
model of form illustrated in Figure~\ref{model}. But we also aim to simultaneously learn a second
posterior distribution $q(z|y,s)$ that enables us to condition each continuous action distribution
on a given scenario. We can accomplish these objectives simultaneously by extending our model with
dual outputs, thereby enabling us to learn two different posterior distributions (encoders). We
illustrate this unified model in Figure~\ref{model_unified}. This unified model contains our
original model in its entirety and is augmented with a second instance of the continuous latent
variable, which we denote $z'$ and a second instance of the trajectory output variable, which we
denote $x'$. We define $p(z'|y) \equiv p(z|y)$ and $p(x'|z') \equiv p(x|z)$, so the same GMM
components and decoder are shared between both branches of the model.
This model reflects the following factorization of generative and variational distributions:
\begin{align}
  \hspace{-1mm}p(x,x',y,z,z'|s)\!&=\!p(y|s)p(z|y)p(x|z)p(z'|y)p(x'|z')\\
  q(y,z,z'|x,s)\!&=\!q(y|x,s)q(z|x)q(z'|y,s).
\end{align}
Following a derivation analogous to~\eqref{elbo_0}-\eqref{elbo}, and using shorthand for each
distribution to simplify notation (e.g., $p(y|s)$ is written $p_y$), we arrive at the following
unified objective:
\begin{align}
  \hspace{-1mm}\mathcal{L}_{\text{unified}} &= \sum_y\int_z\int_{z'} q_y q_z q_{z'}
    \text{ln}\left\{\frac{p_y p_z p_x p_{z'} p_{x'}}{q_y q_z q_{z'}}\right\}dzdz'\\
  \begin{split}
    &\approx \text{ln } p(x|\tilde{z}) + \sum_y q_y\text{ln } p(x'|\tilde{z}')\minus\kl(q_y||p_y)\\
    &\qquad\minus\sum_y q_y\left[\kl(q_z||p_z) + \kl(q_{z'}||p_{z'})\right],
  \end{split}
\end{align}
where $\tilde{z}$ is sampled from $q_z$ and $\tilde{z}'$ is sampled from $q_{z'}$. Inspection of
this expression demonstrates that this model consists exactly of the terms from our original
objective function~\eqref{loss} and the method we proposed to learn a dual encoder in
equation~\eqref{loss_alternate}, seamlessly unifying both objectives.

One important distinction between this unified model and our original model is that our latent
variables no longer follow the mean field approximation due to the dependency of $q(z'|y,s)$ on $y$.
However, we can still follow the same method to derive the following expression for $q(y|x,s)$ in
our unified model (see Appendix~\ref{appendix:derivations}). Again, utilizing shorthand to simplify
notation, we have:
\begin{equation}
  q_y = \frac{p_y\text{exp}(\minus H(q_z,p_z)\minus \kl(q_{z'}||p_{z'}))}
  {\sum_y p_y\text{exp}(\minus H(q_z,p_z)\minus \kl(q_{z'}||p_{z'}))}.
  \label{qy_unified}
\end{equation}
This expression differs from equation~\eqref{qy} only in the appearance of the
$\minus\kl(q_{z'}||p_{z'})$ term, which reflects that the probability of a given data point
belonging to a given cluster also depends on the proximity between the encoding $q(z'|y,s)$ and the
cluster location $p(z'|y)$.

\begin{figure}
  \begin{subfigure}{0.48\linewidth}
    \includegraphics[width=\linewidth,trim=100 220 521 220,clip]{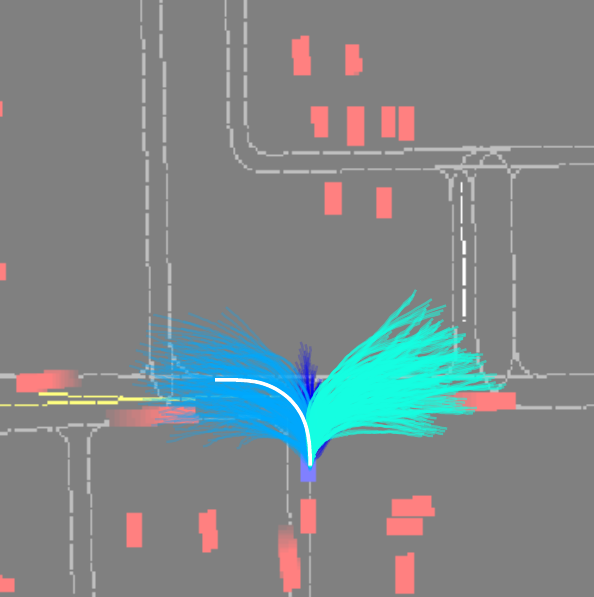}
    \caption{\label{fig:f1}}
  \end{subfigure}
  \hfill
  \begin{subfigure}{0.48\linewidth}
    \includegraphics[width=\linewidth,trim=317 220 304 220,clip]{figures/fig6ab}
    \caption{\label{fig:f2}}
  \end{subfigure}
  \begin{subfigure}{0.48\linewidth}
    \includegraphics[width=\linewidth,trim=150 230 750 235,clip]{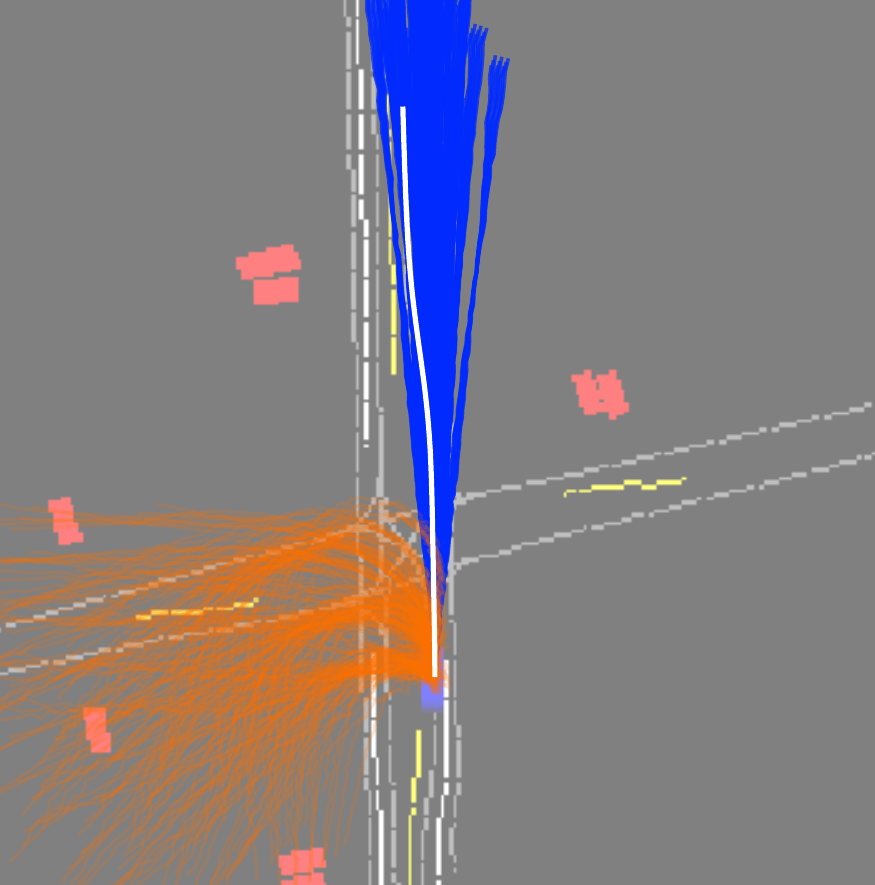}
    \caption{\label{fig:f3}}
  \end{subfigure}
  \hfill
  \begin{subfigure}{0.48\linewidth}
    \includegraphics[width=\linewidth,trim=460 230 440 235,clip]{figures/fig6cd}
    \caption{\label{fig:f4}}
  \end{subfigure}
  \caption{Inferred action distributions produced by the generative distribution $p(z|y)$
  (\subref{fig:f1},\subref{fig:f3}) vs. the variational posterior $q(z|y,s)$
  (\subref{fig:f2},\subref{fig:f4}) (color) and ground truth (white). Only actions with probability
  $p(y|s) > 0.05$ are shown.  Other agents in the scenario are shown in red and lane centerlines are
  illustrated in gray.  See Appendix~\ref{appendix:additional_examples} for additional examples.}
  \label{trajectory_inference}
\end{figure}

Having learned the full unified model, we can make motion predictions by first evaluating $p(y|s)$
to determine the likely action choices, and then evaluating $q(z|y,s)$ for each of the likely action
choices, and decoding those latent distributions through $p(x|z)$ to generate trajectory
distributions.  Figure~\ref{trajectory_inference} illustrates the difference between the generic
action distributions represented by GMM components $p(z|y)$ and the posterior estimates of action
distributions, conditioned on the scenario, given by $q(z|y,s)$. The posterior estimate of the
continuous latent variable $z$ dramatically narrows the distribution from the original GMM component
to a much smaller region of latent space that is likely given the scenario, which, in turn, decodes
to a much more precise distribution of trajectories shown in Figures~\ref{fig:f2} and~\ref{fig:f4}.
We provide additional examples of the unified model in Appendix~\ref{appendix:additional_examples}.

\subsection{Implementation}
\label{sec:implementation}

We trained and tested our models using the Waymo open dataset~\cite{ettinger2021large}. We
implemented a rasterizer to generate inputs with width and height of 224 pixels, three channels of
map information, and additional channels containing the past one second of target and agent states.
The distribution $p(y|s)$ was implemented as a CNN with a ResNet50 feature extraction stage,
although we also observed comparable results with a much smaller CNN. The remaining neural networks
were small feed-forward networks with several hidden layers consisting of 500 nodes each and ReLU
activations.  We defer a thorough exploration of model architectures to future work.

For all results in this paper, we used $z\in\mathbb{R}^5$ and $K=25$. Note that
Figure~\ref{fig:action_distribution} only contains 20 actions due to the fact that five of the
trained actions were never assigned any significant probability in test scenarios and were omitted.
We explored a number of methods for initializing the GMM components in $p(z|y)$, including random
and uniform placement, which function well for very low dimensional continuous latent spaces (e.g.,
$z\in\mathbb{R}^2$).  For higher dimensional latent spaces, we found the most effective means of
initializing $p(z|y)$ was to pre-train $p(x|z)$ and $q(z|x)$ together as a standard variational
autoencoder, encoding some data points into the latent space using $q(z|x)$, and then initializing
GMM components at those locations in latent space. After initialization, all components were jointly
trained to convergence.

\section{Discussion}
\label{sec:discussion}

We have demonstrated in this paper that a latent variable model with both discrete and continuous
latent variables is an effective way to model natural driving data and cluster vehicle behaviors
into discrete actions. Although there are no ``correct'' actions to measure our performance against
in an unsupervised setting, our learned actions are qualitatively reasonable and contain a variety
of turns and straight motions at different speeds. Furthermore, the structure of the model provides
an elegant means of naturally predicting the discrete action probabilities in any given scenario,
which is one of the primary uses of a discrete action representation. As a convenient byproduct, our
model also makes accurate multi-modal motion forecasts with meaningful distributional information
over the discrete action choice as well as the continuous shape of each prediction.

We have proposed two methods for incorporating the scenario $s$ directly into the motion prediction
in Sections~\ref{sec:dual} and~\ref{sec:unified}. One advantage of the first method is that the
structure of the continuous latent space $z$ has no complex dependence on the scenario $s$, which
leads to faster convergence during training, and may result in more definitive distinctions between
the clusters in latent space. On the other hand, the unified model of the second method enables the
sharing of CNN features between $p(y|s)$ and $q(z'|y,s)$, leading to more efficient runtime
performance, and enables the quality of the continuous motion prediction to influence the discrete
action set. An important line of future research will be to quantitatively compare these approaches.
Although motion prediction is not our primary objective, another line of future research will be
optimize and measure the performance of these approaches as motion prediction algorithms.

We have shown that the number of discrete actions needed to describe natural driving can be quite
small ($\approx20$) and is compatible with human-understandable semantic meaning, in contrast with
pure motion prediction methods whose sets of anchor trajectories are often much
larger~\cite{chai2019multipath,phan2020covernet}. It is the continuous latent variable in our model
that enables us to tailor a small number of discrete actions to fit such a wide variety of
continuous vehicle motions. Our model provides a simple means of estimating an effective number of
actions, which is to set a conservatively high value of $K$ and retain only those actions that are
ever assigned a non-negligible probability. One possible direction of future work might be to apply
fully Bayesian methods to explicitly estimate the optimal number of discrete
actions~\cite{10.5555/1162264}.

Finally, although we made no use of manually-defined action categories or labels, it may be useful
to utilize a small set of labeled actions if they are available. Fortunately our models are easily
adapted to use discrete action labels when available. One method is to simply replace $q(y|x,s)$
with the known one-of-$K$ representation of $y$ for the samples whose labels are known. Using
partial labels may help to guide cluster formation for specific actions that are known to exist
while enabling the self-supervised process to discover the additional remaining modes present in the
data.

\clearpage

\bibliographystyle{IEEEtran}
\bibliography{main}

\begin{thebibliography}{10}
\providecommand{\url}[1]{#1}
\csname url@rmstyle\endcsname
\providecommand{\newblock}{\relax}
\providecommand{\bibinfo}[2]{#2}
\providecommand\BIBentrySTDinterwordspacing{\spaceskip=0pt\relax}
\providecommand\BIBentryALTinterwordstretchfactor{4}
\providecommand\BIBentryALTinterwordspacing{\spaceskip=\fontdimen2\font plus
\BIBentryALTinterwordstretchfactor\fontdimen3\font minus
  \fontdimen4\font\relax}
\providecommand\BIBforeignlanguage[2]{{%
\expandafter\ifx\csname l@#1\endcsname\relax
\typeout{** WARNING: IEEEtran.bst: No hyphenation pattern has been}%
\typeout{** loaded for the language `#1'. Using the pattern for}%
\typeout{** the default language instead.}%
\else
\language=\csname l@#1\endcsname
\fi
#2}}

\bibitem{konidaris2019necessity}
G.~Konidaris, ``On the necessity of abstraction,'' \emph{Current opinion in
  behavioral sciences}, vol.~29, pp. 1--7, 2019.

\bibitem{bengio2013representation}
Y.~Bengio, A.~Courville, and P.~Vincent, ``Representation learning: A review
  and new perspectives,'' \emph{IEEE transactions on pattern analysis and
  machine intelligence}, vol.~35, no.~8, pp. 1798--1828, 2013.

\bibitem{Hong2019RulesOT}
J.~Hong, B.~Sapp, and J.~Philbin, ``Rules of the road: Predicting driving
  behavior with a convolutional model of semantic interactions,'' \emph{2019
  IEEE/CVF Conference on Computer Vision and Pattern Recognition (CVPR)}, pp.
  8446--8454, 2019.

\bibitem{cui2019multimodal}
H.~Cui, V.~Radosavljevic, F.-C. Chou, T.-H. Lin, T.~Nguyen, T.-K. Huang,
  J.~Schneider, and N.~Djuric, ``Multimodal trajectory predictions for
  autonomous driving using deep convolutional networks,'' in \emph{2019
  International Conference on Robotics and Automation (ICRA)}.\hskip 1em plus
  0.5em minus 0.4em\relax IEEE, 2019, pp. 2090--2096.

\bibitem{chai2019multipath}
Y.~Chai, B.~Sapp, M.~Bansal, and D.~Anguelov, ``Multi{P}ath: Multiple
  probabilistic anchor trajectory hypotheses for behavior prediction,''
  \emph{arXiv preprint arXiv:1910.05449}, 2019.

\bibitem{phan2020covernet}
T.~Phan-Minh, E.~C. Grigore, F.~A. Boulton, O.~Beijbom, and E.~M. Wolff,
  ``Cover{N}et: Multimodal behavior prediction using trajectory sets,'' in
  \emph{Proceedings of the IEEE/CVF Conference on Computer Vision and Pattern
  Recognition}, 2020, pp. 14\,074--14\,083.

\bibitem{casas2018intentnet}
S.~Casas, W.~Luo, and R.~Urtasun, ``Intent{N}et: Learning to predict intention
  from raw sensor data,'' in \emph{Conference on Robot Learning}.\hskip 1em
  plus 0.5em minus 0.4em\relax PMLR, 2018, pp. 947--956.

\bibitem{lee2017desire}
N.~Lee, W.~Choi, P.~Vernaza, C.~B. Choy, P.~H. Torr, and M.~Chandraker,
  ``Desire: Distant future prediction in dynamic scenes with interacting
  agents,'' in \emph{Proceedings of the IEEE Conference on Computer Vision and
  Pattern Recognition}, 2017, pp. 336--345.

\bibitem{tang2019multiple}
C.~Tang and R.~R. Salakhutdinov, ``Multiple futures prediction,''
  \emph{Advances in Neural Information Processing Systems}, vol.~32, pp.
  15\,424--15\,434, 2019.

\bibitem{casas2020implicit}
S.~Casas, C.~Gulino, S.~Suo, K.~Luo, R.~Liao, and R.~Urtasun, ``Implicit latent
  variable model for scene-consistent motion forecasting,'' in \emph{Computer
  Vision--ECCV 2020: 16th European Conference, Glasgow, UK, August 23--28,
  2020, Proceedings, Part XXIII 16}.\hskip 1em plus 0.5em minus 0.4em\relax
  Springer, 2020, pp. 624--641.

\bibitem{mozaffari2020deep}
S.~Mozaffari, O.~Y. Al-Jarrah, M.~Dianati, P.~Jennings, and A.~Mouzakitis,
  ``Deep learning-based vehicle behavior prediction for autonomous driving
  applications: A review,'' \emph{IEEE Transactions on Intelligent
  Transportation Systems}, 2020.

\bibitem{7995948}
D.~J. Phillips, T.~A. Wheeler, and M.~J. Kochenderfer, ``Generalizable
  intention prediction of human drivers at intersections,'' in \emph{2017 IEEE
  Intelligent Vehicles Symposium (IV)}, 2017, pp. 1665--1670.

\bibitem{deo2018convolutional}
N.~Deo and M.~M. Trivedi, ``Convolutional social pooling for vehicle trajectory
  prediction,'' in \emph{Proceedings of the IEEE Conference on Computer Vision
  and Pattern Recognition Workshops}, 2018, pp. 1468--1476.

\bibitem{zhang2013understanding}
H.~Zhang, A.~Geiger, and R.~Urtasun, ``Understanding high-level semantics by
  modeling traffic patterns,'' in \emph{Proceedings of the IEEE international
  conference on computer vision}, 2013, pp. 3056--3063.

\bibitem{hu2018probabilistic}
Y.~Hu, W.~Zhan, and M.~Tomizuka, ``Probabilistic prediction of vehicle semantic
  intention and motion,'' in \emph{2018 IEEE Intelligent Vehicles Symposium
  (IV)}.\hskip 1em plus 0.5em minus 0.4em\relax IEEE, 2018, pp. 307--313.

\bibitem{ivanovic2020multimodal}
B.~Ivanovic, K.~Leung, E.~Schmerling, and M.~Pavone, ``Multimodal deep
  generative models for trajectory prediction: A conditional variational
  autoencoder approach,'' \emph{IEEE Robotics and Automation Letters}, vol.~6,
  no.~2, pp. 295--302, 2020.

\bibitem{rakos2020compression}
O.~R{\'a}kos, S.~Aradi, T.~B{\'e}csi, and Z.~Szalay, ``Compression of vehicle
  trajectories with a variational autoencoder,'' \emph{Applied Sciences},
  vol.~10, no.~19, p. 6739, 2020.

\bibitem{yuan2019diverse}
Y.~Yuan and K.~Kitani, ``Diverse trajectory forecasting with determinantal
  point processes,'' \emph{arXiv preprint arXiv:1907.04967}, 2019.

\bibitem{huang2020diversitygan}
X.~Huang, S.~G. McGill, J.~A. DeCastro, L.~Fletcher, J.~J. Leonard, B.~C.
  Williams, and G.~Rosman, ``Diversity{GAN}: Diversity-aware vehicle motion
  prediction via latent semantic sampling,'' \emph{IEEE Robotics and Automation
  Letters}, vol.~5, no.~4, pp. 5089--5096, 2020.

\bibitem{kingma2013auto}
D.~P. Kingma and M.~Welling, ``Auto-encoding variational {B}ayes,'' \emph{arXiv
  preprint arXiv:1312.6114}, 2013.

\bibitem{jang2016categorical}
E.~Jang, S.~Gu, and B.~Poole, ``Categorical reparameterization with
  {G}umbel-softmax,'' \emph{arXiv preprint arXiv:1611.01144}, 2016.

\bibitem{joo2020dirichlet}
W.~Joo, W.~Lee, S.~Park, and I.-C. Moon, ``Dirichlet variational autoencoder,''
  \emph{Pattern Recognition}, vol. 107, p. 107514, 2020.

\bibitem{tomczak2018vae}
J.~Tomczak and M.~Welling, ``{VAE} with a {V}amp{P}rior,'' in
  \emph{International Conference on Artificial Intelligence and
  Statistics}.\hskip 1em plus 0.5em minus 0.4em\relax PMLR, 2018, pp.
  1214--1223.

\bibitem{nalisnick2016stick}
E.~Nalisnick and P.~Smyth, ``Stick-breaking variational autoencoders,''
  \emph{arXiv preprint arXiv:1605.06197}, 2016.

\bibitem{rezende2015variational}
D.~Rezende and S.~Mohamed, ``Variational inference with normalizing flows,'' in
  \emph{International conference on machine learning}.\hskip 1em plus 0.5em
  minus 0.4em\relax PMLR, 2015, pp. 1530--1538.

\bibitem{dilokthanakul2016deep}
N.~Dilokthanakul, P.~A. Mediano, M.~Garnelo, M.~C. Lee, H.~Salimbeni,
  K.~Arulkumaran, and M.~Shanahan, ``Deep unsupervised clustering with
  {G}aussian mixture variational autoencoders,'' \emph{arXiv preprint
  arXiv:1611.02648}, 2016.

\bibitem{ijcai2017-273}
\BIBentryALTinterwordspacing
Z.~Jiang, Y.~Zheng, H.~Tan, B.~Tang, and H.~Zhou, ``Variational deep embedding:
  An unsupervised and generative approach to clustering,'' in \emph{Proceedings
  of the Twenty-Sixth International Joint Conference on Artificial
  Intelligence, {IJCAI-17}}, 2017, pp. 1965--1972. [Online]. Available:
  \url{https://doi.org/10.24963/ijcai.2017/273}
\BIBentrySTDinterwordspacing

\bibitem{rao2019continual}
D.~Rao, F.~Visin, A.~Rusu, R.~Pascanu, Y.~W. Teh, and R.~Hadsell, ``Continual
  unsupervised representation learning,'' \emph{Advances in Neural Information
  Processing Systems}, vol.~32, pp. 7647--7657, 2019.

\bibitem{10.5555/1162264}
C.~M. Bishop, \emph{Pattern Recognition and Machine Learning}.\hskip 1em plus
  0.5em minus 0.4em\relax Berlin, Heidelberg: Springer-Verlag, 2006.

\bibitem{ettinger2021large}
S.~Ettinger, S.~Cheng, B.~Caine, C.~Liu, H.~Zhao, S.~Pradhan, Y.~Chai, B.~Sapp,
  C.~R. Qi, Y.~Zhou, \emph{et~al.}, ``Large scale interactive motion
  forecasting for autonomous driving: The {W}aymo open motion dataset,'' in
  \emph{Proceedings of the IEEE/CVF International Conference on Computer
  Vision}, 2021, pp. 9710--9719.

\end{thebibliography}

\appendices
\section{Derivations of $q(y|x,s)$}
\label{appendix:derivations}
To derive $q(y|x,s)$ for the model illustrated in Figure~\ref{model}, we follow the derivation
by~\cite{10.5555/1162264} for the case of the mean field approximation. To simplify notation, we
will use shorthand, e.g., $q_y$ to refer to $q(y|x,s)$, and other distributions will be denoted
similarly. We start by writing the ELBO and rearranging terms to arrive at an expression of KL
divergence between $q_y$ and another distribution. Because our goal is to find the optimal value of
$q_y$ given the current parameters of all other distributions, we will be dropping terms that are
constant with respect to $q_y$.
\begin{align}
  \hspace{-1mm}\mathcal{L} &= \sum_y\int_z q_y q_z \left\{\text{ln}(p_x p_y p_z)\minus\text{ln }q_z\minus\text{ln
  }q_y \right\}dz\\
  &= \sum_y q_y \left\{\int_z q_z\text{ln}(p_x p_y p_z) dz\right\}\minus\sum_y q_y\text{ln }q_y +
  \text{const}\\
  &= \minus\kl(q_y||\tilde{p}),
\end{align}
where we have defined $\text{ln }\tilde{p}$ as:
\begin{equation}
  \text{ln }\tilde{p} \equiv \int_z q_z\text{ln}(p_x p_y p_z) dz + \text{const}.
\end{equation}
$\mathcal{L}$ is optimized with respect to $q_z$ when $\kl(q_y||\tilde{p}) = 0$, which gives:
\begin{equation}
  \text{ln }q_y^{*} = \text{ln }\tilde{p} = \int_z q_z\text{ln}(p_x p_y p_z) dz + \text{const}.
  \label{q_y_int}
\end{equation}
Expanding the terms in integral in equation~\eqref{q_y_int} gives:
\begin{align}
  \text{ln }q_y^{*} &=\nsp\int_z\nsp q_z\text{ln }p_y dz + \nsp\int_z\nsp q_z\text{ln }p_z dz +
  \nsp\int_z\nsp q_z\text{ln }p_x dz + \text{const}\label{q_y_manip}\\
  &=\text{ln }p_y\minus H(q_z,p_z) + \text{const},
\end{align}
where we observe that the third term in equation~\eqref{q_y_manip} does not depend on $q_y$ and so
we combine it with the constant, which we can drop and infer through normalization to give the
result in equation~\eqref{qy}:
\begin{equation}
  q_y^{*} = \frac{p_y\text{exp}(\minus H(q_z,p_z))}{\sum_y p_y\text{exp}(\minus H(q_z,p_z))}.
\end{equation}

The derivation for expression for $q(y|x,s)$ in the unified model illustrated in
Figure~\ref{model_unified} is somewhat more involved due to the fact that the latent variables in
the unified model do not adhere to the mean field approximation. Nevertheless, we can extend the
derivation above to arrive at the expression in equation~\eqref{qy_unified}. Once again, we begin
with the ELBO, collect terms that are constant with respect to $q_y$, and rearrange terms to arrive
at an expression of KL divergence:
\begin{align}
  \begin{split}
    \hspace{-1mm}\mathcal{L}_{\text{unified}} &= \sum_y\int_z\int_{z'} q_y q_z q_{z'} \Bigl\{\text{ln}(p_x p_{x'} p_y p_z p_{z'})\\
    &\qquad\minus\text{ln }q_z\minus\text{ln }q_{z'}\minus\text{ln }q_y \Bigr\}dz'dz
  \end{split}\\
  \begin{split}
    &= \sum_y q_y\Biggl\{\int_z\int_{z'}q_z q_{z'} \text{ln}(p_x p_{x'} p_y p_z p_{z'}) dz'dz\\
    &\qquad\minus\int_{z'}q_{z'}\text{ln }q_{z'}dz'\Biggr\}-\sum_y q_y\text{ln }q_y + \text{const}
  \end{split}\\
  &= \minus\kl(q_y||\tilde{p}_{\text{unified}}),
\end{align}
where we have defined $\text{ln }\tilde{p}_{\text{unified}}$ as:
\begin{equation}
\begin{split}
  \text{ln }\tilde{p}_{\text{unified}} &\equiv  \int_z\int_{z'}q_z q_{z'} \text{ln}(p_x p_{x'} p_y p_z p_{z'}) dz'dz\\
    &\qquad\minus\int_{z'}q_{z'}\text{ln }q_{z'}dz' + \text{const}.
\end{split}\label{p_tilde_unified}
\end{equation}
We observe again that $\mathcal{L}_{\text{unified}}$ is optimized with respect to $q_y$ when
$\kl(q_y||\tilde{p}_{\text{unified}}) = 0$. Expanding and combining terms in
equation~\eqref{p_tilde_unified}, and collecting those that are constant with respect to $q_y$
gives:
\begin{equation}
  \text{ln }q^*_y = \text{ln }p_y \minus H(q_z,p_z) \minus \kl(q_{z'}||p_{z'}) + \text{const}.
\end{equation}
Finally, inferring the constant terms through normalization, we arrive at the result in
equation~\eqref{qy_unified}:
\begin{equation}
  q_y^{*} = \frac{p_y\text{exp}(\minus H(q_z,p_z)\minus \kl(q_{z'}||p_{z'}))}{\sum_y
  p_y\text{exp}(\minus H(q_z,p_z)\minus \kl(q_{z'}||p_{z'}))}.
\end{equation}

\section{Additional Prediction Examples}
\label{appendix:additional_examples}

\begin{figure}
  \begin{subfigure}{0.48\linewidth}
    \includegraphics[width=\linewidth,trim=280 180 270 180,clip]{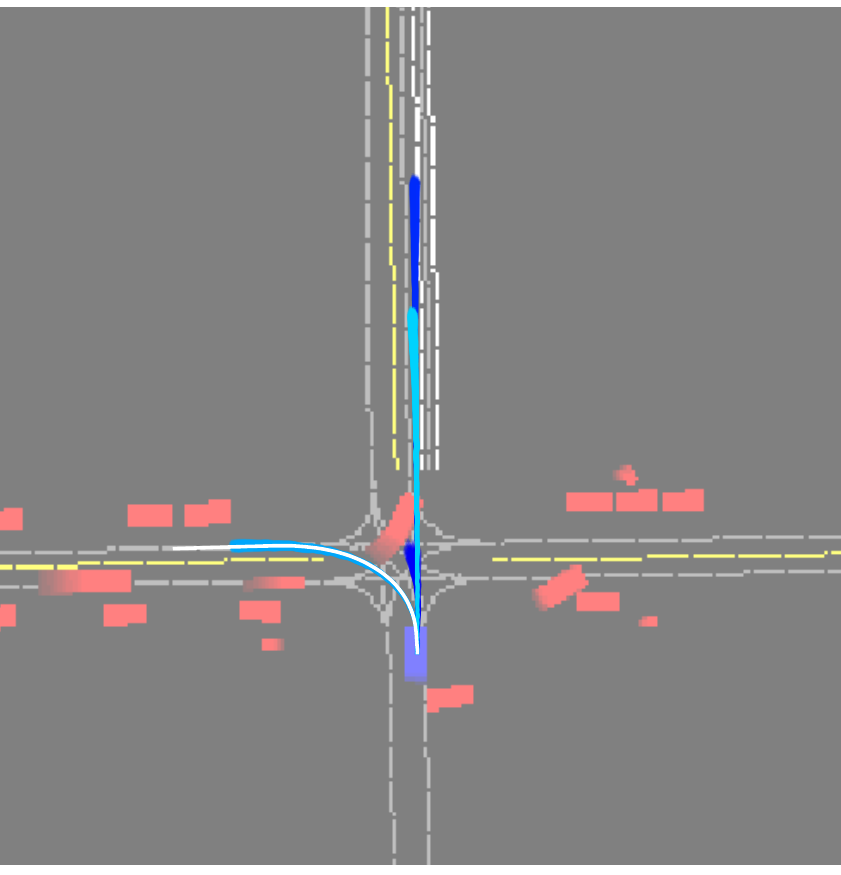}
    \caption{\label{fig:a1}}
  \end{subfigure}
  \hfill
  \begin{subfigure}{0.48\linewidth}
    \includegraphics[width=\linewidth,trim=280 180 270 180,clip]{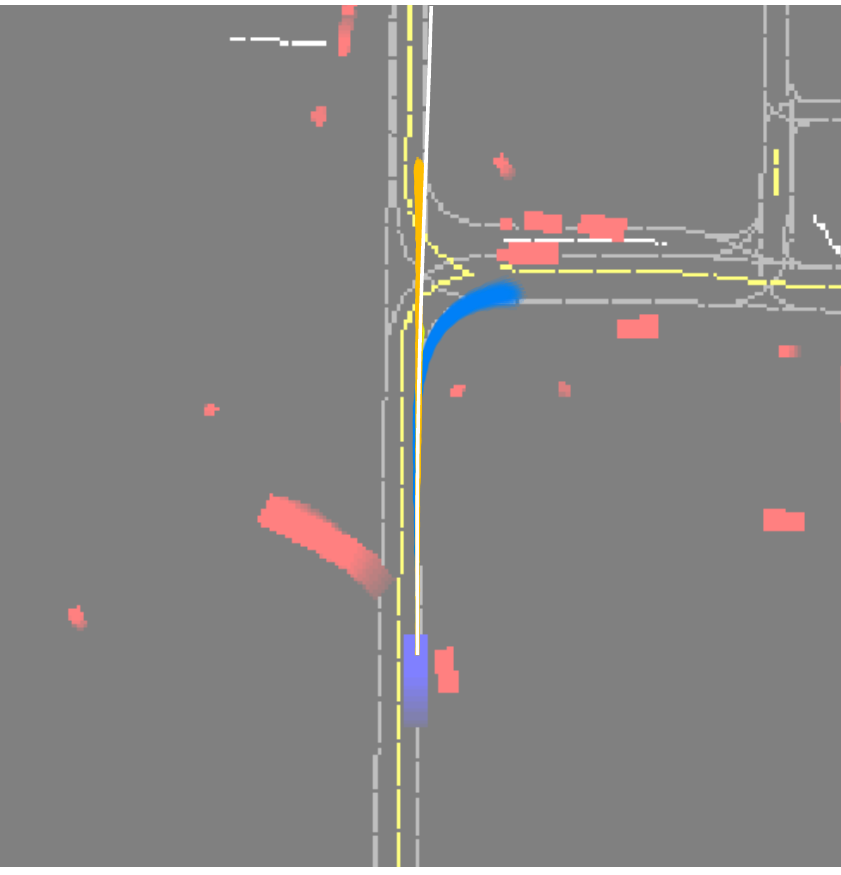}
    \caption{\label{fig:a2}}
  \end{subfigure}
  \begin{subfigure}{0.48\linewidth}
    \includegraphics[width=\linewidth,trim=280 180 270 180,clip]{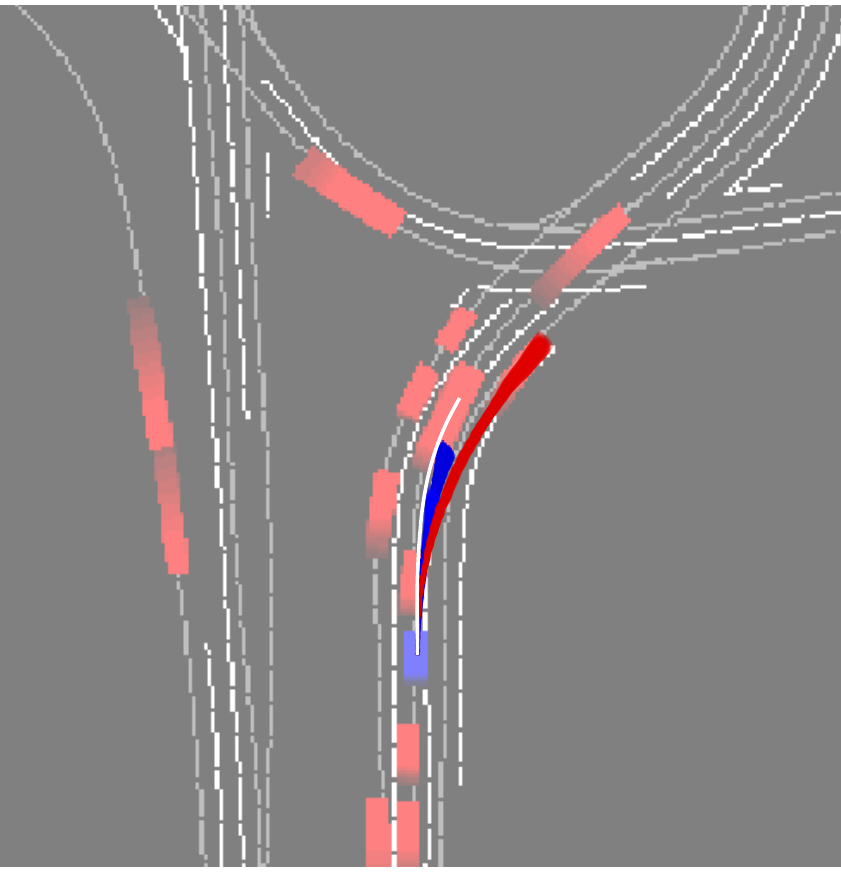}
    \caption{\label{fig:a3}}
  \end{subfigure}
  \hfill
  \begin{subfigure}{0.48\linewidth}
    \includegraphics[width=\linewidth,trim=280 180 270 180,clip]{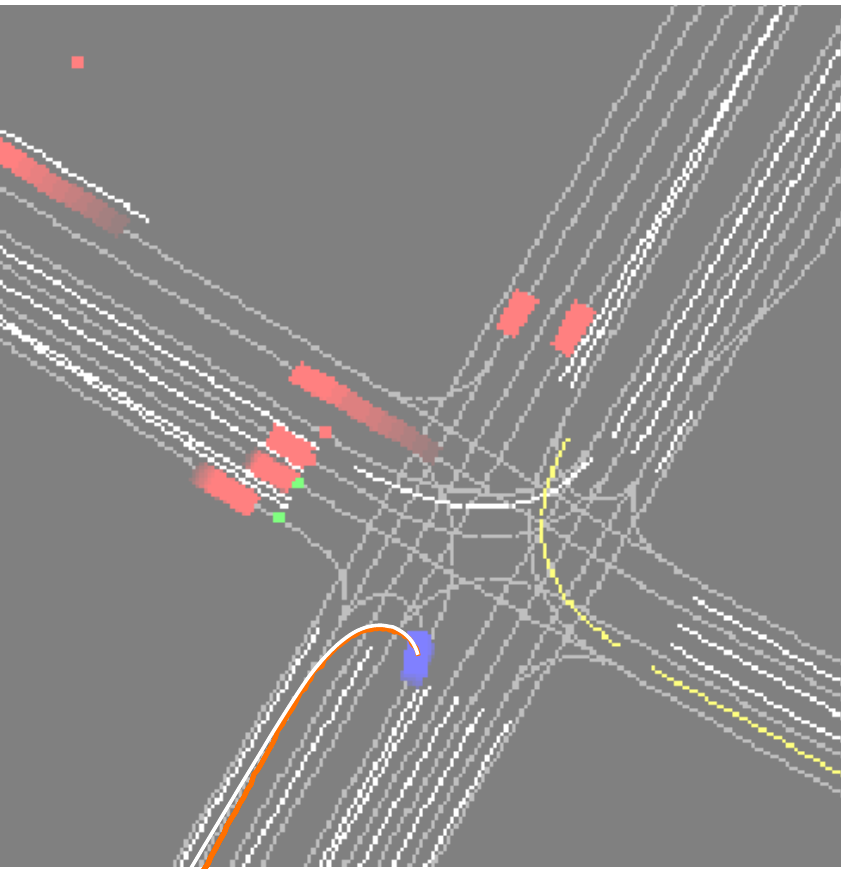}
    \caption{\label{fig:a4}}
  \end{subfigure}
  \begin{subfigure}{0.48\linewidth}
    \includegraphics[width=\linewidth,trim=280 180 270 180,clip]{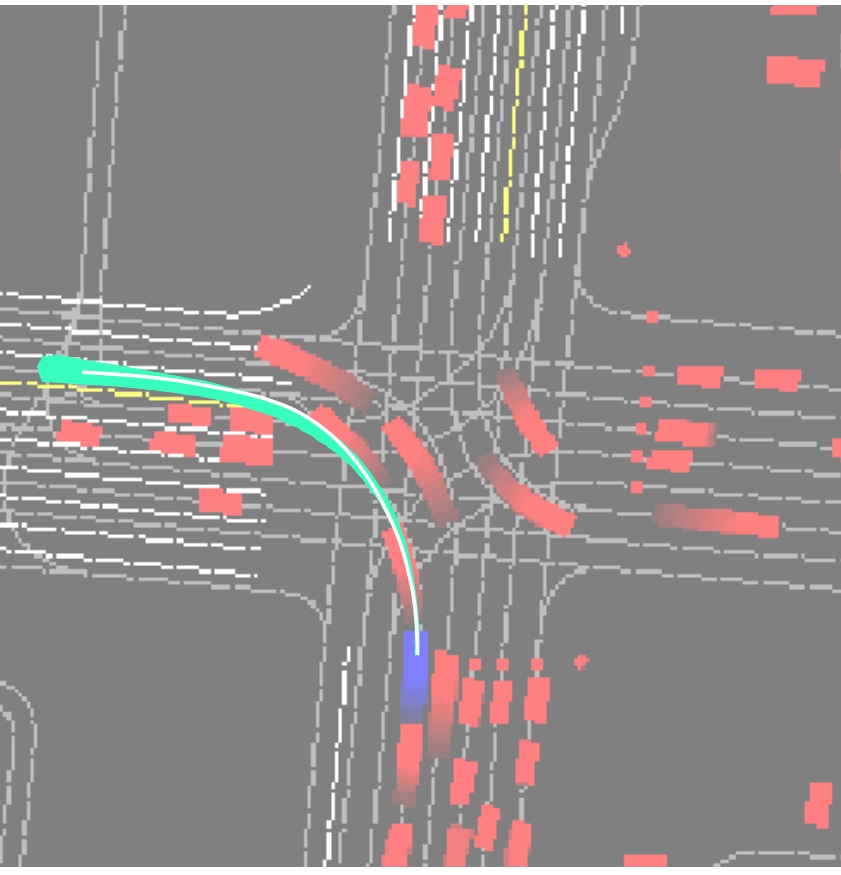}
    \caption{\label{fig:a5}}
  \end{subfigure}
  \hfill
  \begin{subfigure}{0.48\linewidth}
    \includegraphics[width=\linewidth,trim=280 180 270 180,clip]{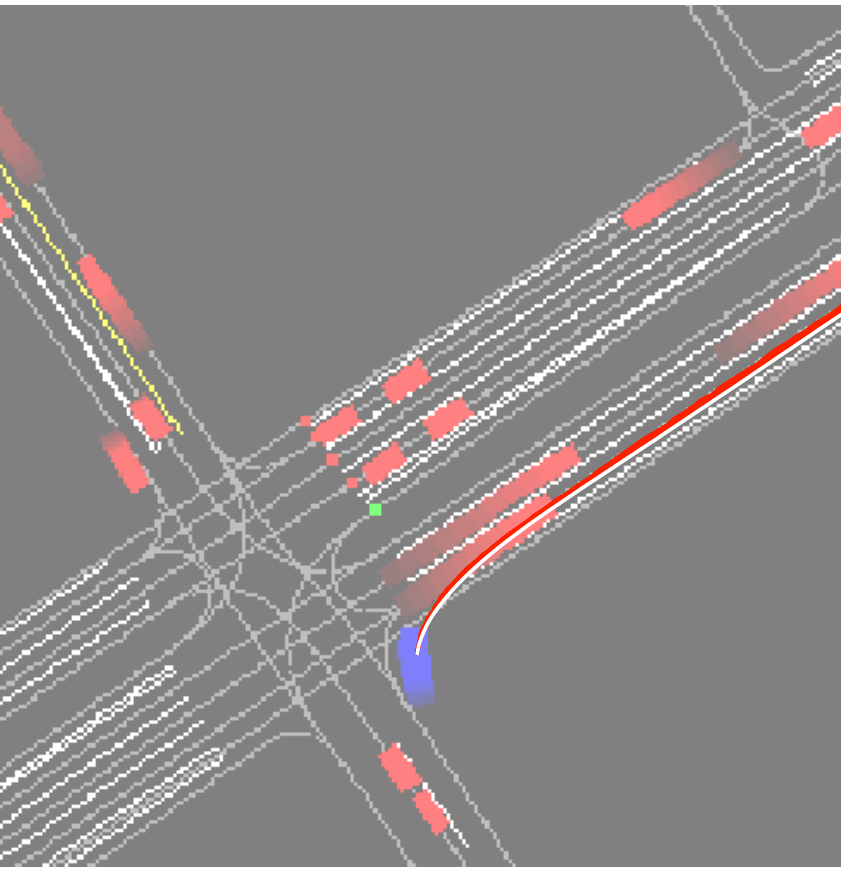}
    \caption{\label{fig:a6}}
  \end{subfigure}
  \begin{subfigure}{0.48\linewidth}
    \includegraphics[width=\linewidth,trim=280 180 270 180,clip]{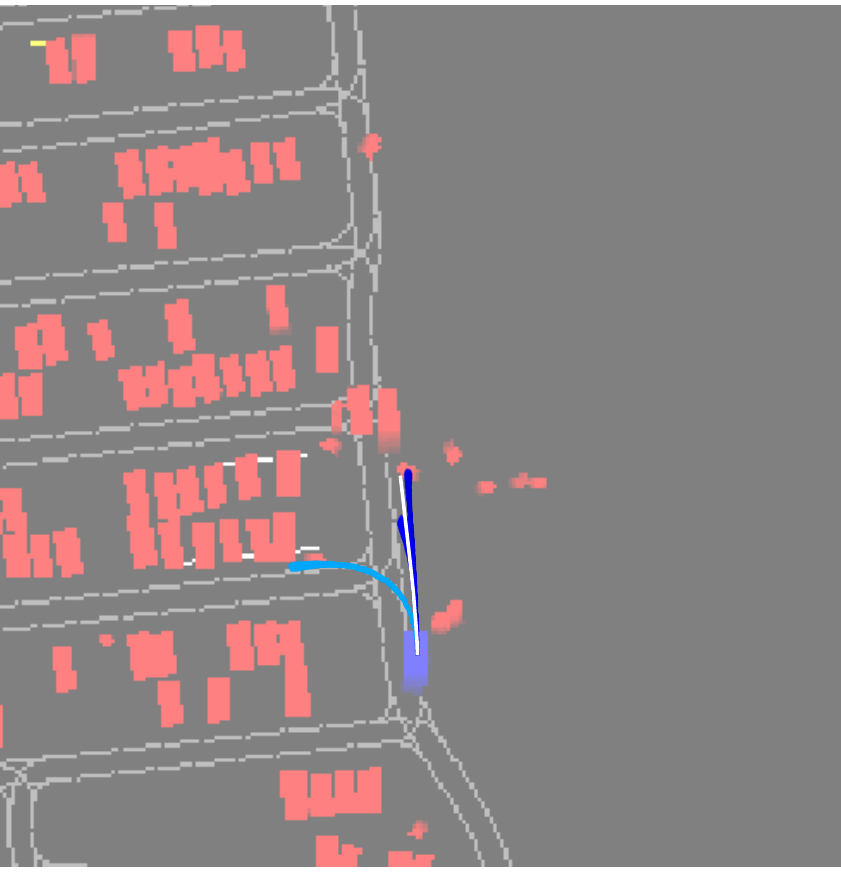}
    \caption{\label{fig:a7}}
  \end{subfigure}
  \hfill
  \begin{subfigure}{0.48\linewidth}
    \includegraphics[width=\linewidth,trim=280 180 270 180,clip]{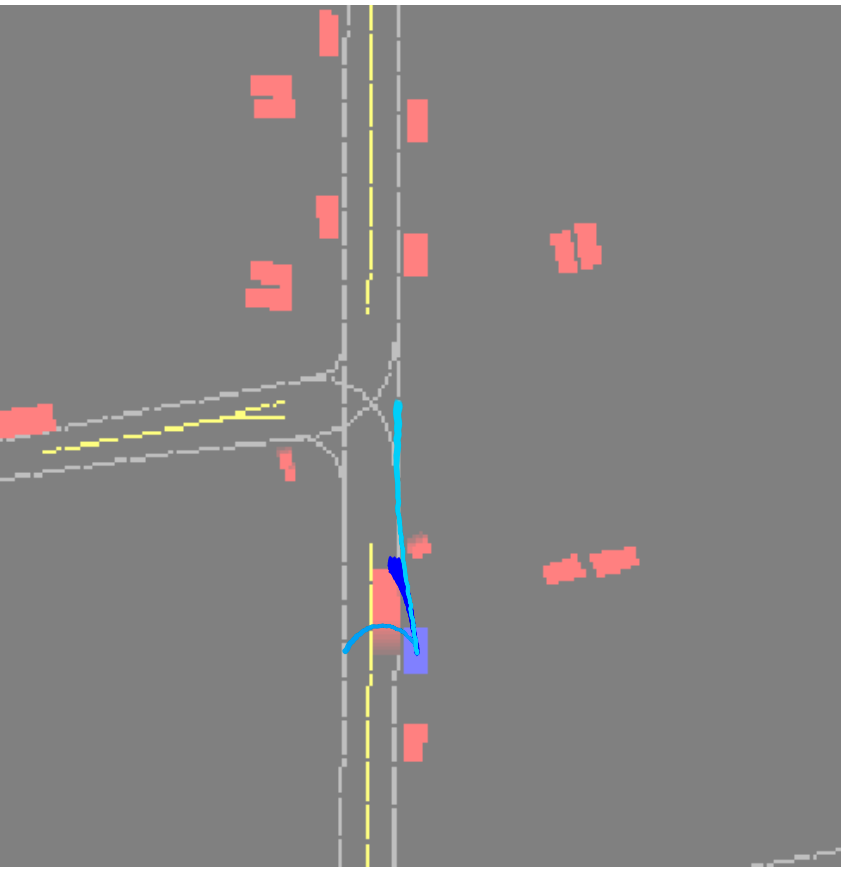}
    \caption{\label{fig:a8}}
  \end{subfigure}
  \caption{Example predictions from the unified model. Colored trajectory distributions are
  predictions of different actions. Ground truth is illustrated in white.
  \subref{fig:a1},\subref{fig:a2} Multi-modal predictions at intersections. \subref{fig:a3} Two
  predictions corresponding to following a leading vehicle and changing lanes to overtake.
  \subref{fig:a4} Performing a U-turn. \subref{fig:a5},\subref{fig:a6} Performing turns from turn
  lanes. \subref{fig:a7} Following pedestrians and vehicles in a parking lot. \subref{fig:a8}
  Pulling slowly out of a parking spot.}
  \label{fig:appendix_examples}
\end{figure}

In this section we provide additional examples of a variety of prediction scenarios to demonstrate
the versatility of our approach. In some cases multiple options are feasible and likely, which often
occurs at intersections and in unstructured situations. In other cases, only a single action is
assigned any significant probability, for example in a turn lane when no other actions are feasible,
or in the middle of an obvious maneuver such as a U-turn. The images in
Figure~\ref{fig:appendix_examples} illustrate examples of these cases.

\end{document}